\documentclass[acmsmall,screen,9pt]{acmart}

\AtBeginDocument{%
  }

\usepackage{graphicx}                    
\usepackage{newtxmath}                   
\usepackage{multirow}                    
\usepackage{xcolor}                      
\usepackage{textcomp}                    
\usepackage{booktabs}                    
\usepackage{algorithm}                   
\usepackage{algpseudocode}               
\usepackage{listings}                    
\usepackage{url}                         
\usepackage{enumerate}
\usepackage{algorithm}
\usepackage{algpseudocode}
\usepackage{tikz}
\usepackage{float}
\usetikzlibrary{matrix,arrows.meta, positioning}
\usepackage{xspace}
\newcommand{\tensorcomp}{ToxiTenCompl\xspace}
\newcommand{\toxicomp}{ToxiCompl\xspace}

\begin{document}

\title[]{Completion of the DrugMatrix Toxicogenomics Database using 3-Dimensional Tensors}

\author{Tan Nguyen}
\authornote{Both authors contributed equally to this research.}
\email{nguyed21@unlv.nevada.edu}
\orcid{0009-0002-2045-5945}
\author{Guojing Cong}
\authornotemark[1]
\email{congg@ornl.gov}
\affiliation{%
  \institution{Oak Ridge National Laboratory}
  \city{Oak Ridge}
  \state{TN}
  \country{USA}
}

\renewcommand{\shortauthors}{}
\begin{abstract}
	We explore applying a tensor completion approach to complete the DrugMatrix toxicogenomics dataset. Our hypothesis is that by preserving the 3-dimensional structure of the data, which comprises tissue, treatment, and transcriptomic measurements, and by leveraging a machine learning formulation, our approach will improve upon prior state-of-the-art results.  Our results demonstrate that the new tensor-based method more accurately reflects the original data distribution and effectively captures organ-specific variability. The proposed tensor-based methodology achieved lower mean squared errors and mean absolute errors compared to both conventional Canonical Polyadic decomposition and 2-dimensional matrix factorization methods. In addition, our non-negative tensor completion implementation reveals relationships among tissues. Our findings not only complete the world's largest in-vivo toxicogenomics database with improved accuracy but also offer a promising methodology for future studies of drugs that may cross species barriers, for example, from rats to humans.
\end{abstract}

\maketitle

\section{Introduction}

It is crucial to understand the risk of intended and adverse effects of compounds in drug discovery. In addition to in vitro and in vivo approaches, toxicity datasets such as Drugmatrix\cite{Chen_Zhang_Borlak_Tong_2012}, TG-GATEs \cite{Igarashi_Nakatsu_Yamashita_Ono_Ohno_Urushidani_Yamada_2014}, DRUG-Seq\cite{Ye2018DRUG}, LINCS\cite{Subramanian2017Next}, CMAP\cite{Lamb2006Connectivity}, and sci-Plex\cite{Srivatsan2020Massively}  are used to profile the toxicity of compounds \cite{liu2020tensor}.

The DrugMatrix toxicogenomics database has become a key resource for studying molecular and apical toxicity profiles of short-term \textit{in vivo} rat studies \cite{Ganter2006DrugMatrix, Auerbach2010}. It contains gene-expression data (from multiple microarray platforms) and traditional endpoints such as histopathology, clinical chemistry, and hematology for hundreds of compounds, but organized in a sparse, largely incomplete fashion \cite{Cong2024ToxiCompl}. Conventionally, these data are viewed as a \emph{2D matrix}, with rows denoting probes or endpoints and columns representing (compound--dose--time) treatments \cite{Auerbach2010}, and recent efforts (e.g., ToxiCompl) have used matrix-factorization techniques to fill in the missing entries \cite{Cong2024ToxiCompl}.

Despite these advances, the 2-dimensional viewpoint can be restrictive because, in reality, the toxicogenomics data are naturally \emph{multi-dimensional}. For instance, each probe measurement is made with a \emph{compound}, a \emph{dose}, a \emph{duration}, for a \emph{gene} within a \emph{tissue}, suggesting a natural \emph{3D or 4D tensor} representation \cite{Kolda2009TensorReview,Chi2013TensorCompletion,Auerbach2010}. By capturing higher-dimensional relationships among these variables, tensor factorization or tensor completion approaches may exploit correlation structures across multiple modes more effectively than 2D matrix completion methods \cite{liu2020tensor}. Tensor completion may offer several advantages in modeling toxicogenomic data. First, it enables the preservation of multi-mode interactions by simultaneously modeling compound–dose–time relationships and tissue-specific effects. This approach captures multi-modal patterns that can reveal biologically meaningful gene-expression variations across tissues and treatment conditions \cite{Kolda2009TensorReview,Chi2013TensorCompletion}. Second, tensor methods may improved imputation accuracy under data sparsity \cite{Kolda2009TensorReview, liu2020tensor} because they model a richer set of relationships than 2D matrix completion. Third, tensor models offer enhanced interpretability, particularly when using non-negative tensor factorization (NTF)~\cite{Zhao2024-ew} . NTF can yield additive, coherent factor matrices that reflect underlying biological processes.  Finally, tensor completion helps reduce bias from central-tendency smoothing, where imputed values are pulled toward a global average, especially under high sparsity, a common issue in 2D factorization techniques~\cite{Chi2013TensorCompletion}.

However, to our knowledge, no prior studies have applied a 3D or higher-dimensional completion framework to the DrugMatrix or similar datasets. The restriction of 2D modeling potentially leaves crucial biological relationships under-exploited. Tensor completion and factorization approaches, such as Canonical Polyadic (CP), Tucker Decomposition (TD), and tensor Singular Value Decomposition (TSVD), abound in the literature, each offering certain advantages for modeling multi-dimensional relationships among the data. For instance, CP explicitly factorizes each mode with independent factor matrices \cite{Kolda2009TensorReview}, facilitating clear interpretations of interactions among tissues, treatments and transcripts. TD introduces additional flexibility through a core tensor that encodes correlations withtin tensor slices \cite{Kolda2009TensorReview}. TSVD leverages structural correlations within tensor slices, ensuring computational efficiency and robustness against noise and sparsity \cite{Kolda2009TensorReview, Kilmer2011}. 

Many existing tensor completion algorithms, such as the convex relaxation method by Yuan and Zhang~\cite{Yuan2016TensorRecovery}, the semidefinite programming (SDP)-based method by Barak and Moitra ~\cite{Barak2016TensorSDP}, and the alternating minimization framework with strong orthogonality assumptions proposed by Jain and Oh~\cite{Jain2014Alternating}, are either highly heuristic with slow convergence, based on large SDPs that are impractical to run in practice, or rely on unrealistic assumptions such as requiring the factor matrices to be nearly orthogonal. We propose a new 3D tensor completion algorithm we call \tensorcomp that is formulated in a machine learning context—solved via gradient descent rather than the standard alternating minimization paradigm. In contrast to other implementations, \tensorcomp with DrugMatrix converges much faster and with superior accuracy. It also better preserves meaningful rare signals in the data. Our non-negative version of \tensorcomp yields interpretable factors for tissues that are conducive to studying relationships among them.

Our contributions are as follows:
\begin{itemize}
\item We propose a 3D tensor formulation of DrugMatrix, and our completion algorithm, \tensorcomp, produces more accurate predictions than SOTA baselines.
\item \tensorcomp per iteration is much faster than the prior 2D matrix completion implementation, \emph{ToxCompl}, due to reduced matrix factor sizes.
\item The non-negative implementation of \tensorcomp is able to produce factor matrices that may be used to study relationships among tissues
\end{itemize}


\section{Data and Prior Approaches}
\label{s:data-approach}
The DrugMatrix dataset has approx. $n_1=193{,}000$ rows and $n_2=3{,}000$ columns, amounting to a theoretical maximum of around 580 millions entries. The data comprise histopathology, clinical chemistry, hematology, and gene-expression measurements (on both the Codelink ``RU1'' and Affymetrix ``RG230'' platforms). Each column in the data set corresponds to a unique treatment group (e.g., chemical--dose--duration), and each row corresponds to a different endpoint (e.g., a specific gene prob, a histopathology score, a clinical chemistry measurement, and so forth).

Table~\ref{tab:data-comp} illustrates the distribution of the \emph{gene-expression} data across two platforms (RU1, RG230) and eight tissues (livers, kidney, heart, bone marrow, brain, intestine, spleen, skeletal muscle). Notably, liver (LI) and kidney (KI) together make up the largest fraction of observed entries, reflecting their high relevance in toxicology. In contrast, tissues like brain (BR) and intestine (IN) are studied much less often, with corresponding fewer measurements.

\begin{table*}[h]
	\centering
	\begin{tabular}{| c | c | c | c | c | c | c | c | c | c |}
		\hline
		\textbf{RU1} & \textbf{RG230} & \textbf{LI} & \textbf{KI} & \textbf{HE} & \textbf{BM} & \textbf{BR} & \textbf{IN} & \textbf{SP} & \textbf{SM} \\
		\hline
		32.6M        & 39.4M          & 34.5M       & 19.0M       & 11.8M       & 2.7M        & 0.55M       & 0.17M       & 1.5M        & 1.5M        \\
		\hline
	\end{tabular}
	\caption{Data of different categories across gene-expression platforms (RU1 vs.\ RG230) and organs:
		LI = liver, KI = kidney, HE = heart, BM = bone marrow, BR = brain, IN = intestine, SP = spleen, SM = skeletal muscle. 
		Each cell shows the count of observed gene-expression measurements (in millions), illustrating
	significant differences in coverage across tissues.}
	\label{tab:data-comp}
\end{table*}

Table~\ref{tab:category-dist} shows that the vast majority of measured fold‐changes in DrugMatrix fall into Category 0—values near zero—comprising 91.94 \% of all entries. Moderate downregulation (Category –1) and upregulation (Category 1) account for 4.09 \% and 3.88 \% of the data, respectively, while extreme downregulation (Category –2) and upregulation (Category 2) are both very rare (0.03 \% and 0.036 \%). This extreme skew toward near‐zero changes highlights the challenge of imputing a dataset dominated by small or negligible effects and underscores.

\begin{table}[h]
  \centering
  \begin{tabular}{|c|c|c|c|c|}
    \hline
    \textbf{Category\,–2} & \textbf{Category\,–1} & \textbf{Category\,0} & \textbf{Category\,1} & \textbf{Category\,2} \\
    \hline
    0.03\% & 4.09\% & 91.94\% & 3.88\% & 0.036\% \\
    \hline
  \end{tabular}
  \caption{Distribution of categorical bins in the original DrugMatrix data and in the test predictions made by the original (ToxiCompl) model.}
  \label{tab:category-dist}
\end{table}

\subsection{PCA, Autoencoders, and Generic Matrix Factorization} 
    Principal component analysis (PCA) is a traditional strategy to map the data into a lower-dimensional space, seeking the directions of greatest variance in the original high-dimensional matrix~\cite{Jolliffe2016PCAReview}. While PCA often helps reduce noise and highlight key variation, it can only capture \emph{linear} relationships and can struggle with the heavily skewed distribution of expression values in DrugMatrix (where roughly 92\% of entries lie in a near-zero region~\cite{Hasan2019Cluster}).  
    In contrast, \emph{autoencoders} can model nonlinear patterns via encoder-decoder architectures~\cite{Bank2020Autoencoders}, and there exist variants such as variational autoencoder (VAE)~\cite{Kingma2019VAE}, adversarial autoencoder (AAE)~\cite{Makhzani2015Adversarial}, and Siamese autoencoders~\cite{Baier2023Siamese} that can potentially preserve more subtle features in toxicogenomics data. However, applying these methods directly to the large, skewed DrugMatrix dataset has typically produced only modest success in clustering tasks, with many rare but toxicologically significant signals underrepresented~\cite{Auerbach2010}.
    
    Generic matrix factorization approximates the data with the product of smaller factor matrices. This can be coupled with specialized loss functions or attention mechanisms to account for DrugMatrix’s skewed data distribution. Despite improved preservation of rare or extreme gene-expression shifts, such 2D factorization still “flattens” the inherent multi-way structure (e.g., ignoring time/dose/tissue axes) and has not always yielded well-defined clusters when validated with standard algorithms like K-means or DBSCAN~\cite{Hasan2019Cluster}.

\subsection{Toxicogenomis-aware 2D Completion}

\toxicomp \cite{Cong2024ToxiCompl} is a recent toxicogenomics-ware implementation that completes the DrugMatrix with high accuracy. Its predictions are validated from both the machine learning perspective and the toxicology perspective.  Compared with generic 2D matrix completion, \toxicomp formulates the completion as a machine learning problem and pays special attention to preserving the rare signals and matching the distribution of the predicted data with that of the training data. \toxicomp introduces a robust loss function accounting for the skewed distribution of gene-expression values. Additionally, it incorporates side information, such as known compound-target interactions or toxicological annotaitons, in an effort to guide the factorization process. These advances helps increase the accuracy of DrugMatrix completion. We use \toxicomp as one of the baselines in our study.

\subsection{Generic 3D Tensor Completion Approaches}

Generic 3D tensor completion approaches leverage tensor algebra and multilinear analysis to impute missing entries effectively. Song et al. \cite{song2019tensor} categorizes these traditional tensor completion methodologies into three primary groups:

\begin{enumerate}
    \item \textbf{Decomposition-Based Approaches:}
    These methods utilize tensor decompositions, primarily CANDECOMP/PARAFAC (CP) and Tucker decomposition. The CP decomposition factorizes a tensor into a sum of rank-one tensors:    
    \[
    \mathcal{X} \approx \sum_{r=1}^{R} \mathbf{a}_r^{(1)} \circ \mathbf{a}_r^{(2)} \circ \dots \circ \mathbf{a}_r^{(N)}
    \]
    
    where \(\mathbf{a}_r^{(n)}\) represents factor matrices. Tucker decomposition generalizes this by including a core tensor multiplied by factor matrices across each mode:
    
    \[
    \mathcal{X} \approx \mathcal{C} \times_1 \mathbf{A}^{(1)} \times_2 \mathbf{A}^{(2)} \times_3 \dots \times_N \mathbf{A}^{(N)}
    \]
    These methods exploit low-rank structures, capturing higher-order correlations across tensor modes, making them particularly suitable for structured, sparse biological data.
    
    \item \textbf{Trace Norm-Based Approaches:}
    
    Trace norm approaches generalize matrix nuclear norms to tensors, providing convex optimization frameworks:
    
    \[
    \min_{\mathcal{X}} \sum_{n=1}^{N} \alpha_n \|X_{(n)}\|_* \quad \text{s.t.} \quad \mathcal{X}_{\Omega} = \mathcal{T}_{\Omega} + \mathcal{E}_{\Omega}
    \]
    
    where \(\|X_{(n)}\|_*\) denotes the nuclear norm of tensor unfolding along the \(n\)-th mode. The trace norm encourages low-rank solutions, making this approach effective at regularizing and completing tensors when the data are sparse or incomplete.
    
    \item \textbf{Other Variants:}
    
    Additional specialized variants have emerged addressing specific conditions or constraints:
    
    \begin{itemize}
        \item \textit{Non-negative constrained approaches:} enforce non-negativity to enhance interpretability.
        \item \textit{Robust tensor completion:} incorporates robust principal component analysis to manage corrupted data effectively.
        \item \textit{Riemannian optimization:} optimizes tensor completion on smooth manifolds defined by fixed-rank constraints, which has shown efficiency in computations.
    \end{itemize}

\end{enumerate}

A notable advancement in tensor completion methodologies is proposed by Liu and Moitra \cite{liu2020tensor}, who introduced a novel variant of alternating minimization named \textit{Kronecker Alternating Minimization}. Traditional alternating minimization methods update factor matrices iteratively by fixing two factor matrices and solving for the thrid, typically using the Khatri-Rao product. Liu and Moitra observed that this conventional approach often truggles, particularly when factors are highly correlated, leading to slow convergence and suboptimal solutions.

To overcome these challenges, Liu and Moitra's method replaces the Khatri-Rao product with the Kronecker product, effectively expanding the problem from $r$ rank-one components to $r^2$ rank-one components. This adaptation significantly enhances convergence behavior by stabilizing updates and reducing the likelihood of the algorithm becoming trapped inpoor local minima.

Mathemtically, their updated optimization step becomes:
\[
\{\mathbf{z}_{i,j}\} = \arg\min_{\mathbf{z}_{i,j}} \left\| \left(\mathcal{T} - \sum_{1 \leq i,j \leq r} \hat{\mathbf{x}}_i \otimes \hat{\mathbf{y}}_j \otimes \mathbf{z}_{i,j}\right)\bigg|_{\Omega} \right\|_F^2,
\]
where the set $\{\mathbf{z}_{i,j}\}$ consists of vectors to be optimized, and $\Omega$ represents observed tensor entries. After solving the expanded least squares problem, the resulting factor vectors $\{\mathbf{z}_{i,j}\}$ are reduced back to the original rank via a singular value decomposition (SVD)-based approximation step.

The main theoretical contribution from Liu and Moitra is that this method enjoys strong theoretical guarantees under mild assumptions: robust linear independence of tensor factors; and incoherence of factor matrices, ensuring a balanced distribution of tensor information.
Their algorithm achieves provable convergence at a linear rate to the true tensor, even when factors are highly correlated. Moreover, Liu and Moitra provide rigorous bounds on computational complexity, demonstrating their approach can operate efficiently at near-linear runtime relative to the number of observed entries. Unfortunately, leveraging Liu and Moitra's implementation currently remains challenging due to complexity in practical reproducibility, and no publicly available software implementations have been released, hindering broader adoption.

As a representative of generic 3D tensor completion algorithms, we use the CP implementation in PyTen. PyTen is a python package containing the state-of-the-art tensor decomposition and completion algorithms for “filling in the gaps”' of recovering high-order tensor-structured datasets characterized by noisy and missing information. 


\section{A New 3D Approach for Completing DrugMatrix}


Our tensor completion algorithm, \tensorcomp,  consists of several stages designed to handle the complexity and multi-dimensional structure of the DrugMatrix toxicogenomics dataset. This structured pipeline, detailed in Figure~\ref{fig:workflow}, integrates data restructuring, preprocessing, factorization/completion, optimization, validation, and reconstruction.  In contrast to generic tensor completion algorithms, \tensorcomp formulate the completion in a machine learning setting where the factor matrices are treated as weights to be learned according to observed endpoints in DrugMatrix.  This is similar to the \toxicomp approach except it is in a 3D tensor setting. \tensorcomp also introduces an attention mechanism that improves the prediction accuracy.  The details of the components in the pipeline are described as follows. 

\begin{figure}[]
    \resizebox{0.5\textwidth}{!}{%
    \begin{tikzpicture}[>=Stealth,thick]
        \tikzset{
            tensor/.style={rectangle,draw,minimum width=2.5cm,minimum height=1.3cm,align=center,fill=orange!15},
            matrix2d/.style={rectangle,draw,minimum width=2cm,minimum height=1.3cm,align=center,fill=blue!10},
            latent/.style={rectangle,draw,rounded corners,minimum width=2cm,minimum height=1.2cm,align=center,fill=green!15},
            process/.style={rectangle,draw,rounded corners,minimum width=2cm,minimum height=1cm,align=center,fill=gray!20},
            arrow/.style={->,thick}
        }
        \node[matrix2d] (raw2d) {Raw Data\\(2D Matrix)};
                
        \node[process,right=1.0cm of raw2d] (mapping) {Tensor Construction\\(Index mapping)};
                
        \node[tensor,right=1.0cm of mapping] (tensor3d) {$\mathcal{X}_{r,u,i}$\\3D Tensor};
        \node[tensor,right=1.0cm of tensor3d] (holdout) {Holdout cold start};
                
        \node[process,below=0.5cm of tensor3d] (factorization) {Tensor\\Factorization};
                
        \node[latent,below left=0.5cm and 1.0cm of factorization] (latentR) {Tissue Factors\\$R\times k$};
        \node[latent,below=0.8cm of factorization] (latentU) {Transcript Factors\\$U\times k$};
        \node[latent,below right=0.5cm and 1.0cm of factorization] (latentI) {Treatment Factors\\$I\times            k$};
        \node[process, below=0.8cm of latentU] (modeling) {Modeling\\[0.3em] $\hat{y}_{r,u,i} =          
                \sum_{d=1}^{k} r_r^{(d)} \cdot u_u^{(d)} \cdot i_i^{(d)}$};
            \node[process, below=0.8cm of modeling] (validation) {Validation\\[0.3em]
                $\mathcal{L}_{\text{MSE}} = \frac{1}{N}\sum_{(r,u,i)}(y_{r,u,i}-\hat{y}_{r,u,i})^2$};
            \node[tensor,below = 0.8cm of validation] (completed3d) {Completed                                                        Tensor\\$\hat{\mathcal{X}}_{r,u,i}$};
            
            \draw[arrow] (raw2d) -- (mapping);
            \draw[arrow] (mapping) -- (tensor3d);
            \draw[arrow] (tensor3d) -- (holdout);
            \draw[arrow] (tensor3d) -- (factorization);
                    
            \draw[arrow] (factorization) -- (latentR);
            \draw[arrow] (factorization) -- (latentU);
            \draw[arrow] (factorization) -- (latentI);
                    
            \draw[arrow] (latentR.south) |- (modeling.west);
            \draw[arrow] (latentU.south) -- (modeling.north);
            \draw[arrow] (latentI.south) |- (modeling.east);
            
            \draw[arrow] (modeling.south) -- (validation.north);
            \draw[arrow] (holdout.east) -- ++(1.0,0) |- (validation.east);
            \draw[arrow] (validation.south) -- (completed3d.north);
                    
            \draw[arrow] (factorization.west) -- ++(-5.0,0) |- (completed3d.west) node[midway, below]{Reconstruction};
        \end{tikzpicture}
        }
    \caption{Workflow of \tensorcomp}
    \label{fig:workflow}
\end{figure}

\subsection{Conversion from 2D Matrix to 3D Tensor}

The DrugMatrix data are represented as a 2D matrix, where rows denote transcriptomic measurements (gene expression levels) and columns correspond to distinct treatment conditions characterized by compound, dosage, and exposure duration.  The same genes appear in different tissues as described in Section~\ref{s:data-approach}. Note the same genes in different tissues may have very different expresssion levels to the same treatment. To leverage and model the relationship among the same genes in different tissues, we explicitly restructure the data into a 3D tensor.

Formally, we define this tensor as \(\mathcal{X}\in\mathbb{R}^{R\times U\times I}\), where \(R\) denotes the number of distinct tissues (or tissue-platform combinations), \(U\) represents the number of unique genes, and \(I\) corresponds to the number of unique treatment conditions. Each entry \(\mathcal{X}_{r,u,i}\) captures the expression level of transcript \(u\) in tissue \(r\) under treatment \(i\). The transformation from 2D matrix to 3D tensor involves precise index mapping from original DrugMatrix (denoted as $\mathcal{G}$)  to their corresponding tensor $\mathcal{X}$:
\[\mathcal{X}_{r,u,i} \leftarrow \mathcal{G}_{r\times 8 \times |U|+u,i}\]


\subsection{Data Preprocessing and Cold-Start Holdout}

Before tensor factorization, rigorous preprocessing steps ensure the quality and reliability of the data. Firstly, potential outliers are identified and removed through statistical thresholding based on z-scores calculated for each transcript. Subsequently, normalization techniques, such as Min-Max scaling or StandardScaler, are applied to harmonize gene expression values and stabilize variance across different experimental conditions.

Additionally, we implement a cold-start holdout strategy wherein a randomly selected subset of data entries is withheld from the training process to serve as a robust validation set. This method ensures that our tensor completion model is rigorously evaluated for its predictive generalization capabilities.

\subsection{Attention augmented 3D Tensor Complation}

The core of our approach is the our formulation of a  tensor factorization method. This approach factorizes the constructed tensor into three separate latent factor matrices corresponding to tissues, genes, and treatments, respectively: $\mathbf{R}\in\mathbb{R}^{R\times k}$, $\mathbf{U}\in\mathbb{R}^{U\times k}$, $\mathbf{I}\in\mathbb{R}^{I\times k}$
where \(k\) is the rank determining the dimensionality of the latent representation and controls the complexity of the model. Each observed value in the tensor \(\mathcal{X}\) is modeled through a multilinear predictive function that combines these latent factors. Formally, predicted values \(\hat{y}_{r,u,i}\) are computed as follows:

\[
\hat{y}_{r,u,i}=\sum_{d=1}^{k}(r_{r}^{(d)}\cdot u_{u}^{(d)}\cdot i_{i}^{(d)})
\]

An attention mechanism is included to dynamically adjust the contribution of individual latent components, which is particularly useful in capturing complex interactions. The attention-weighted prediction is given by:

\[
a_{r,u,i}^{(d)}=\text{Softmax}(r_{r,a}^{(d)}\cdot u_{u,a}^{(d)}\cdot i_{i,a}^{(d)})
\]

leading to the enhanced predictive model:

\[
\hat{y}_{r,u,i}=\sum_{d=1}^{k}(r_{r}^{(d)}\cdot u_{u}^{(d)}\cdot i_{i}^{(d)}\cdot a_{r,u,i}^{(d)})+b_u+b_i+b
\]

where \(b_u\), \(b_i\), and \(b\) represent gene-specific, treatment-specific, and global bias terms, respectively. The latent factors can also be enriched with biological and chemical side information via embedding Multi-Layer Perceptrons (MLPs), further enhancing interpretability.

\subsection{Custom Loss Functions and Metrics}

The Model parameters are optimized using stochastic gradient descent (SGD) with the objective of minimizing predictive errors. A natural loss function is mean squared error (MSE) between prediction and ground-truth:
\[
\mathcal{L}_{\text{MSE}}=\frac{1}{N}\sum_{(r,u,i)}(y_{r,u,i}-\hat{y}_{r,u,i})^2
\]
Considering the skewed data distribution in DrugMatrix and hence the tensor $\mathcal{X}$, we adopt a revised, weighted MSE loss, assigning higher weights \(w_{r,u,i}\) to over- and under-expressed genes:
\[
\mathcal{L}_{\text{weighted}}=\frac{1}{N}\sum_{(r,u,i)}w_{r,u,i}(y_{r,u,i}-\hat{y}_{r,u,i})^2-\lambda\cdot\text{Var}(\hat{y})
\]
Here $\text{Var}(\hat{y})$ measures the deviation of ground truth from the mean values of gene expression.  Weighted MSE loss balances overall prediction performance and performance for rare signals. 


In addition to MAE and MSE, we use two other metrics on the holdout set for validation. The first is weighted mean absolute error (WeightedMAE) that prioritizes biologically critical observations through weighting $\text{WeightedMAE}=\frac{1}{N}\sum_{i=1}^{N}w_i|y_i-\hat{y}_i| $. The second is maximum absolute error (MaxAE) that captures worst-case deviations: $\text{MaxAE}=\max_i|y_i-\hat{y}_i|$.








\subsection{Tensor Reconstruction}

  The matrix factors that yield the best performance on the holdout set is used to reconstruct the entire tensor \(\hat{\mathcal{X}}\), which may then be used for various downstream analysis such as pathway analysis and drug effect discovery.  This reconstructed dataset enables meaningful biological interpretation and facilitates research into toxicity mechanisms and drug effects. We also map \(\hat{\mathcal{X}}\) back to the original DrugMatrix format, \(\hat{\mathcal{G}}\) for comparison. 





\section{Results}

In this section, we present detailed findings comparing the \tensorcomp tensor completion method with the \toxicomp matrix completion approach and the CP algorithm. As the values in DrugMatrix is skewed, we evaluate the overall MAE, the MAE for rare signals (over- and under-expressed genes), and the distribution of the values. Ideally the predicted endpoints should have similar data distribution as the input data. We aim to validate our hypotheses that 1. modeling the data in 3D tensor can boost prediction performance, 2. a machine learning formulation can bring performance advantages over classical approaches, and 3. our attention augmented mechanism will produce data distributions that better fit the input. 

\subsection{\tensorcomp vs. \toxicomp vs. CP }

\begin{figure}[h!]
    \centering
  \includegraphics[width=0.45\linewidth]{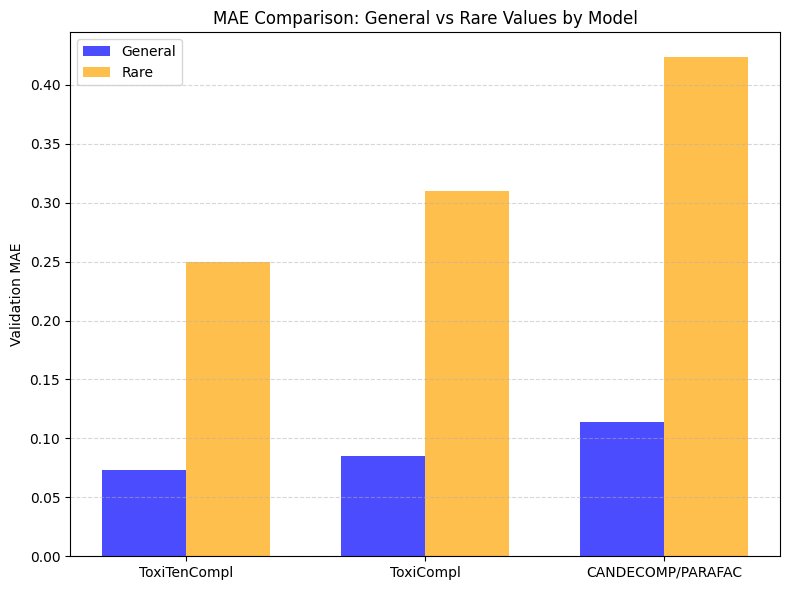} 
  \caption{Overall MAE and MAE for rare signals: a comparison of \tensorcomp against the two baselines}
    \label{fig:comp-3}
\end{figure}

Figure~\ref{fig:comp-3} shows the overall MAE and MAE for rare signals for \tensorcomp, \toxicomp, and CP.  Among the three, \tensorcomp achieves the lowest MAE in both scenarios, demonstrating the advantage of both the 3D tensor setting and the attention mechanism; both \tensorcomp and \toxicomp perform better than CP, demonstrating the advantage of a machine learning setting over traditional approaches.

\subsection{Detailed Comparison of \tensorcomp and \toxicomp}

We investigate the training of \tensorcomp and \toxicomp. We show the evolution of MAE, MSE, Weighted MAE, and MaxAE with the number of training epochs. We use a factor size $k=300$ in both implementations. The optimizer is ADAM with learning rate $lr=0.001$ and weigh decay $5e-4$. We train for 100 epochs with a grace period of 5 epochs. 

\begin{figure}[h!]
    \centering
    \includegraphics[width=0.45\linewidth]{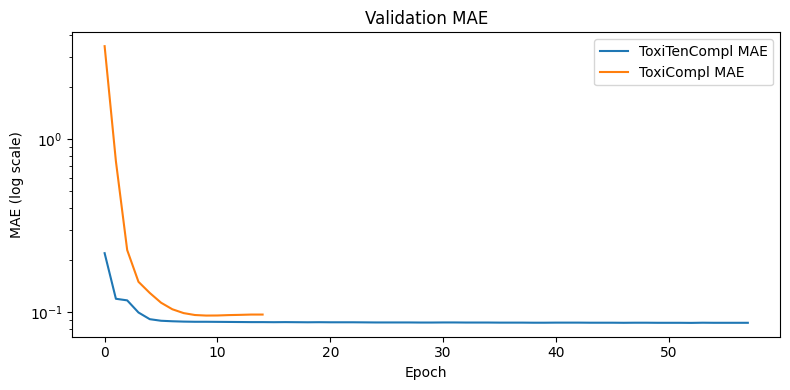} \hspace{0.5cm} \includegraphics[width=0.45\linewidth]{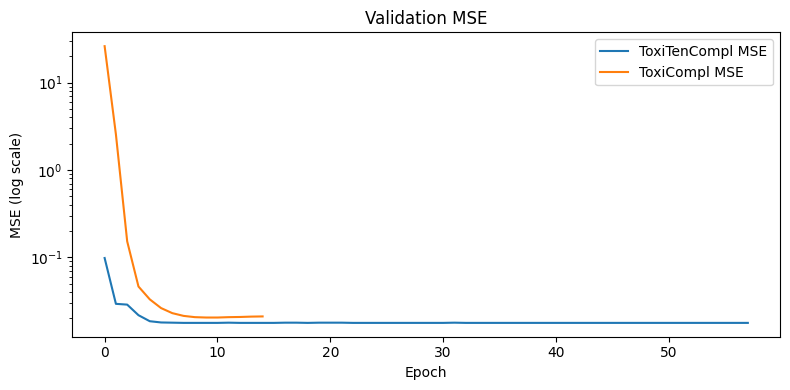}
    \includegraphics[width=0.45\linewidth]{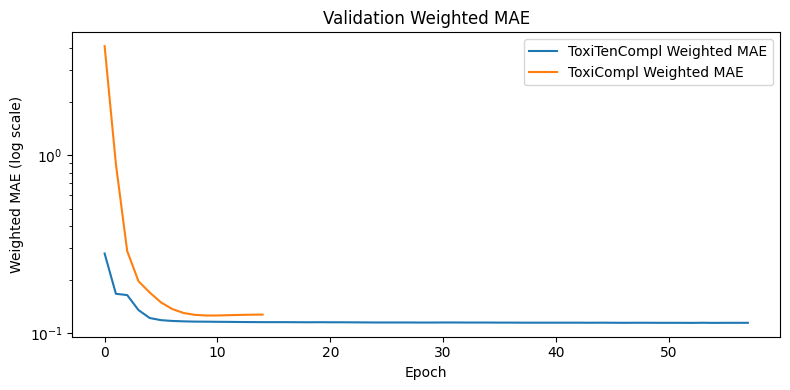} \hspace{0.5cm} \includegraphics[width=0.45\linewidth]{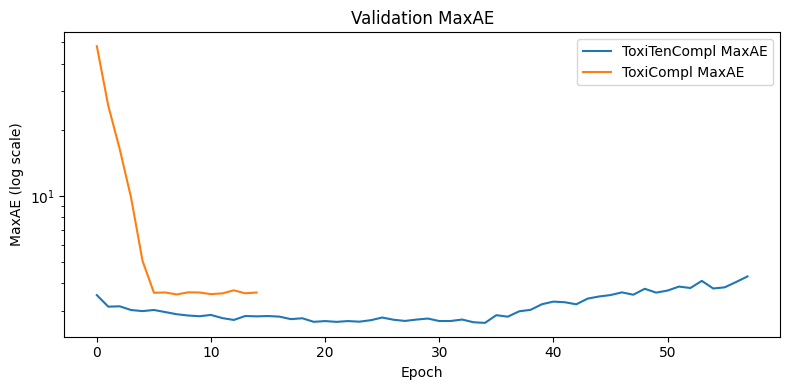}
    \caption{Validation MSE \& MAE comparison between \tensorcomp and \toxicomp. \tensorcomp achieved substantially lower MSE and continued training effectively across epochs, while \toxicomp plateaued and stopped improving after epoch 14.}
    \label{fig:mse_training}
\end{figure}

The plots in Figure~\ref{fig:mse_training} show the performance of \tensorcomp compared to \toxicomp. The validation MSE, MAE, and weighted MAE  curves clearly indicate that \tensorcomp performs better than \toxicomp. 

\toxicomp training halts at epoch 14 due to the lack of further improvement, indicating limitations in its capacity to reduce error metrics further. In contrast, \tensorcomp continues improving to beyond 50 epochs. The MaxAE curve fluctuates among there is a tug of war to balance the maximum prediction error (mostly for rare signals)  and the general weighted error for all. 

\subsection{Prediction Distribution: \tensorcomp vs. \toxicomp}

\begin{figure}[h!]
    \centering
    \includegraphics[width=0.6\linewidth]{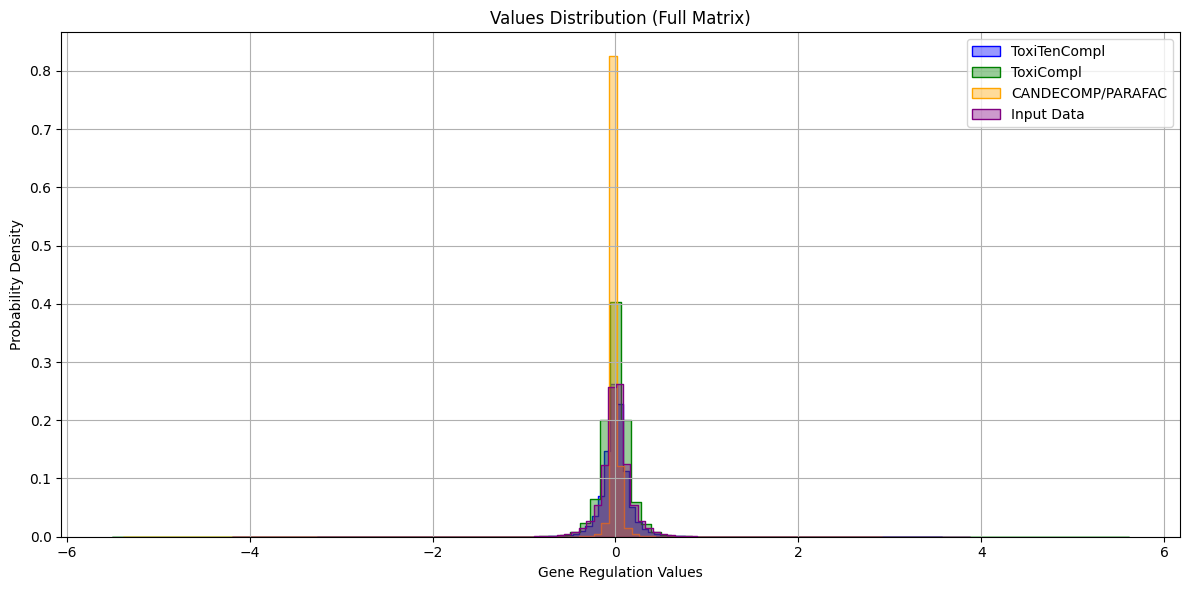}
    \caption{Comparison of prediction distributions from \toxicomp and \tensorcomp. \tensorcomp effectively avoids the excessive central tendency observed in \toxicomp predictions, thereby preserving biologically relevant variability}
    \label{fig:distribution}
\end{figure}

Figure~\ref{fig:distribution} compares the distributions of predicted value generated by \tensorcomp and \toxicomp. \tensorcomp notably avoids the tendency to predict the mean of data (central tendency), a characteristic of the \toxicomp predictions. Central tendency diminishes the model's sensitivity to biologically meaningful over- and under-expressed genes. In contrast, \tensorcomp better preserves these significant variations, confirming it captures more nuanced biological responses in gene expression. In the figure we also include the distribution of the predictions for CP, which is extremely concentrated at the mean.  This confirms our hypothesis that machine learning (to be exact, deep learning) based methods are more adaptive to the input data (a reason could be that in deep learning formulations, we include non-linear activation functions). 

\begin{figure}[h!]
    \centering
    \includegraphics[width=0.6\linewidth]{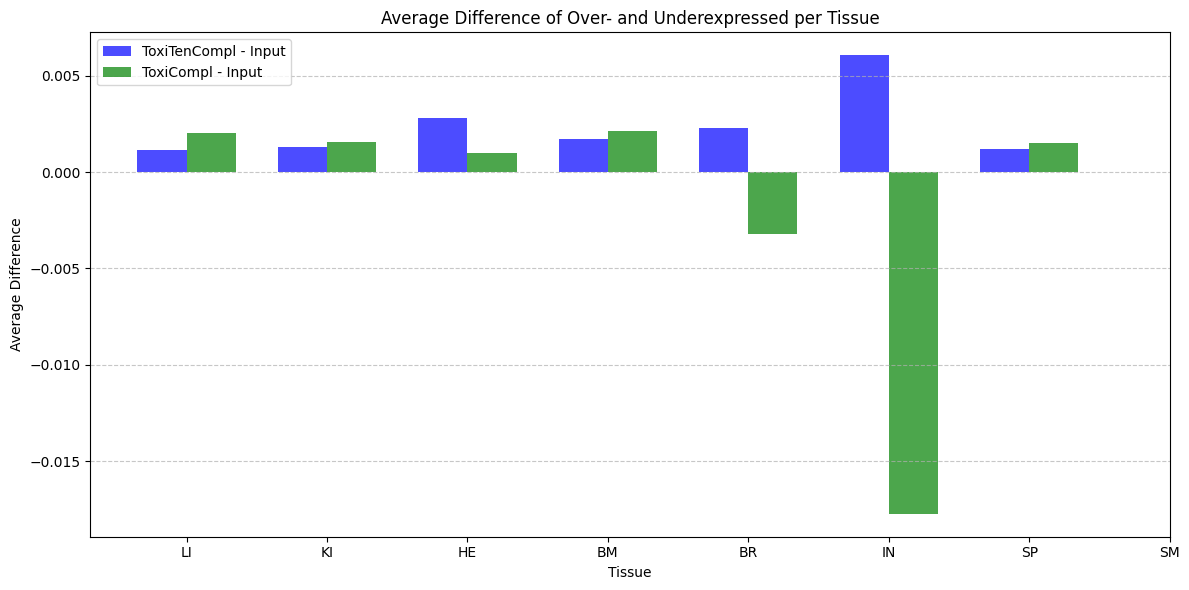}
    \caption{Average difference between \tensorcomp vs. \toxicomp predictions and original input data, by tissue. \tensorcomp predictions consistently show closer alignment with input data distribution, except for Heart (HE)}
    \label{fig:tissue_comparison}
\end{figure}

Figure~\ref{fig:tissue_comparison} shows the average difference of predicted expression values from actual input data, broken down by tissue type. \tensorcomp predictions demonstrate closer alignment with the original input data compared to \toxicomp across nearly all tissues, with the exception of brain tissue.

Quantitatively, \tensorcomp’s deviation from the input data points is consistently smaller than \toxicomp's, highlighting its ability to accurately recover biologically meaningful patterns from sparse and incomplete data. The divergence observed in heart warrants further investigation but broadly does not undermine the overall superior performance of \tensorcomp.

Overall, these results clearly indicate that \tensorcomp not only optimizes training more effectively but also produces biologically relevant predictions with greater accuracy and sensitivity than traditional \toxicomp matrix completion methods.

\subsection{Insights Gained for Tissues}

\begin{figure}[htpb]
    \centering
\includegraphics[width=0.5\linewidth]{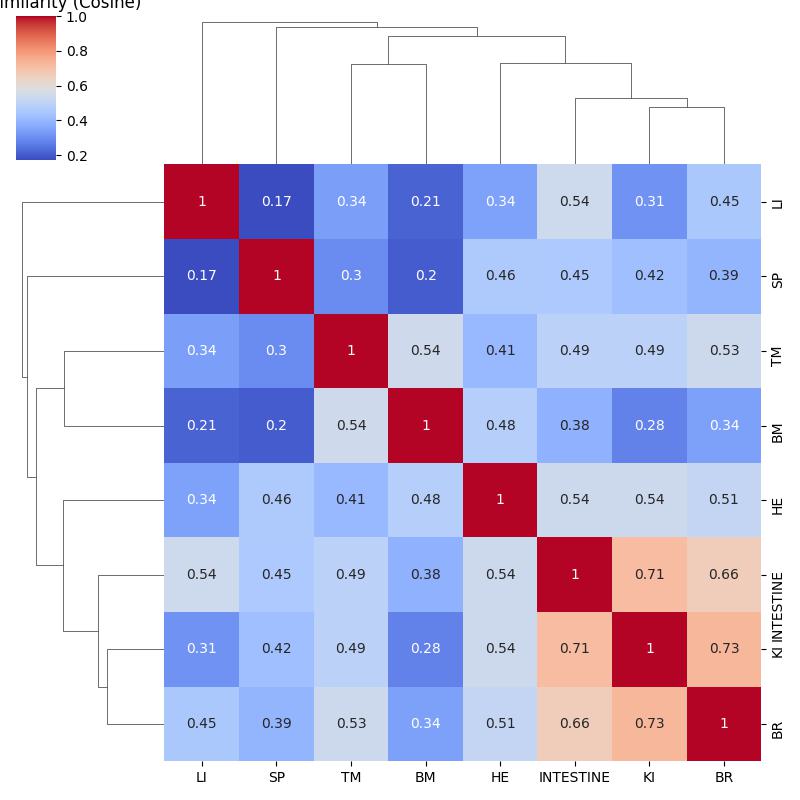}
    \caption{Clustering of the tissue factors}
    \label{fig:insights}
\end{figure}

Modeling DrugMatrix in 3D tensors can also reveal insights about the tissues.  We used a non-negative factorization variant of \tensorcomp to complete the tensor. In this implementation, we first convert the tensor to be non-negative by adding -min($\mathcal{X}$), and then enforce the weights to be non-negative during gradient updates.  Figure~\ref{fig:insights} shows the clustering based on Cosine similarities among the factors for different tissues. 
In Figure~\ref{fig:insights} we observe higher similarities between factors for certain tissues, for example, BM, KI, and INTESTINE, than others. Further analysis on clustering of the genes may reveal patterns of expression networks. Such tissue-to-tissue correlations may reflect shared biological processes or experimental conditions, but further experimental validation and biological interpretation are necessary to elucidate their exact biological relevance.


\section{Discussion \& Conclusion}

In this study we show that a new tensor-based completion method (\tensorcomp)  is superior to the two baselines, \toxicomp and CP, with lower validation errors. This confirms our hypothesis that modeling the DrugMatrix as a 3D tensor can better capture the underlying structure in the data. Specifically, the \tensorcomp predictions avoid the excessive central tendency observed in CP, preserving the biologically relevant variability within gene expression data. The closer alignment of \tensorcomp predictions to input data distributions for most tissues (excluding BR) reinforces this conclusion.



Our study underscores the potential advantage of tensor-based completion methods in a deep learning formulation for biological datasets. By explicitly modeling higher-order interactions among tissues, treatments, and gene expressions, \tensorcomp better preserves biologically meaningful variability compared to approaches with 2D matrices. In addition, it is also possible to discover more insights along each dimensions, for example, the tissue dimension, with potential impact on biology, toxicology, and medicine. 


Several limitations of our approach must be acknowledged and point to future research directions. Primarily, the predictions need to be further validated from a biological or toxicological perspective. Future studies should seek to incorporate experimental validations or biological benchmarks (e.g., independent datasets, pathway analyses, or functional validations) to strengthen the biological interpretability and reliability of tensor-based completions. Additionally, due to lack of measured data for tissues such as BR and IN, \tensorcomp does not perform as well for them as for tissues with more data such as LI and KI. In future work we plan to leverage the relationships of treatments to model the data in even higher-dimensional tensors than 3D and incorporate domain-specific side information (e.g., known molecular interactions, chemical structures, pathway annotations). Such an effort can also improve interpretability of the predictions.



\section{Ackowledgement}
This material is based upon work supported in part by the
U.S. Department of Energy, Office of Science, Office of Advanced
Scientific Computing Research's's Applied Mathematics Competitive Portfolios program, under contract number
DE-AC05-00OR22725. 
 
This research used resources of the Compute and Data Environment for
Science (CADES) at the Oak Ridge National Laboratory, which is
supported by the Office of Science of the U.S. Department of Energy
under Contract No. DE-AC05-00OR22725.

\bibliographystyle{ACM-Reference-Format}
\bibliography{main}
\end{document}